# Krysalis Hand: A Lightweight, High-Payload, 18-DoF Anthropomorphic End-Effector for Robotic Learning and Dexterous Manipulation

Al Arsh Basheer, Justin Chang, Yuyang Chen, David Kim, and Iman Soltani (Member, IEEE)
Department of Mechanical and Aerospace Engineering
University of California-Davis

*Abstract*—Existing multi-finger robotic hands face several limitations, including excessive weight, mechanical complexity, high cost, and constraints in both payload capacity and degrees of freedom (DoF). These challenges hinder their wide adoption, especially when paired with collaborative robotic arms with limited payload capacity. To address these challenges, we present Krysalis Hand, a five-finger robotic end-effector that combines a lightweight design, high payload capacity, and a high number of degrees of freedom (DoF) to enable dexterous manipulation in both industrial and research settings. This design integrates the actuators within the hand while maintaining an anthropomorphic form. Each finger joint features a self-locking mechanism that allows the hand to sustain large external forces without active motor engagement. This approach shifts the payload limitation from the motor strength to the mechanical strength of the hand, allowing the use of smaller, more cost-effective motors. With 18 DoF and weighing only 790 grams, the Krysalis Hand delivers an active squeezing force of 10 N per finger and supports a passive payload capacity exceeding 10 lbs. These characteristics make Krysalis Hand one of the lightest, strongest, and most dexterous robotic end-effectors of its kind. Experimental evaluations validate its ability to perform intricate manipulation tasks and handle heavy payloads, underscoring its potential for industrial applications as well as academic research. All code related to the Krysalis Hand, including control and teleoperation, is available on our GitHub repository: https://github.com/Soltanilara/Krysalis_Hand.

## I. INTRODUCTION

THE rise of automation in recent decades has fundamentally transformed modern manufacturing, delivering greater efficiency, reduced costs, and increased adaptability [1]. However, due to software complexity, hardware constraints, and limited adaptability, assembly floors have been the least beneficiaries of automation. The technological lag in assembly automation, partly rooted in Moravec's paradox, stems primarily from the technical complexity of even the simplest tasks, such as threading a wire through a hole or connecting an electrical plug [2], let alone assembling intricate parts. Such tasks rely on a complex combination of human dexterity and the ability to process various sensory inputs, including tactile, visual, and auditory feedback [3], [4], which has evolved in humans over millions of years. On the software front, with recent advances in machine learning, the assimilation of large volumes of multi-modal sensory data and the generation of high-dimensional actions is now more feasible than ever before [5], [6]. However, progress on the hardware front has not kept pace. Most industrial robotic grippers are designed primarily for repetitive and simple tasks such as pick-and-place operations, with limited object manipulation capabilities, making them unsuitable for assembly or other more complex tasks encountered on assembly floors [3]. With commonly available two or three-finger grippers [7], adapting to new tasks remains challenging, requiring new research and development for each new automation task [8]. This further limits the advancement in automation by making it economically unjustifiable, especially when dealing with small batches or custom products [9]. Other more complex designs that offer high DoFs are heavy [10] and often prohibitively expensive, limiting access to many research laboratories and hence slowing innovation in this domain. As such, despite the common misconception of extensive automation of assembly activities in large industries, such as automotive, most of these tasks are still manually executed by human operators.

There is a pressing need for affordable robotic hands that meet the requirements of dexterous manipulation. Such a design must be lightweight to ensure compatibility with a wide range of lower cost robotic arms without significantly reducing their payload capacity. It should offer sufficient DoFs to perform complex tasks while handling large payloads, which is often encountered on assembly floors. Furthermore, since production parts are typically designed with the dexterity capabilities of human workers in mind [11], robotic end-effectors must incorporate anthropomorphic features to maximize their effectiveness [12]. The need for anthropomorphism is further motivated by recent advances in machine learning-based techniques such as Learning from Demonstration (LfD) [6], [13]–[20] and Explainable AI (XAI) [21], [22]. LfD enables the quick and cost-effective implementation of automation by allowing robots to learn directly from human demonstrations [23], while XAI improves transparency by allowing robots to explain their actions and manipulation strategies [21]. Anthropomorphic designs facilitate these methods by aligning the action space of the human and the robot, thereby simplifying demonstration, execution, and interpretation [24], [25]. Finally, affordability remains a critical factor in the advancement of this field, ensuring that advanced robotic hands are accessible to more research laboratories and industries. This, in turn, fuels innovation in dexterous manipulation, enhancing manufacturing automation while extending robotic applications to service industries and households.

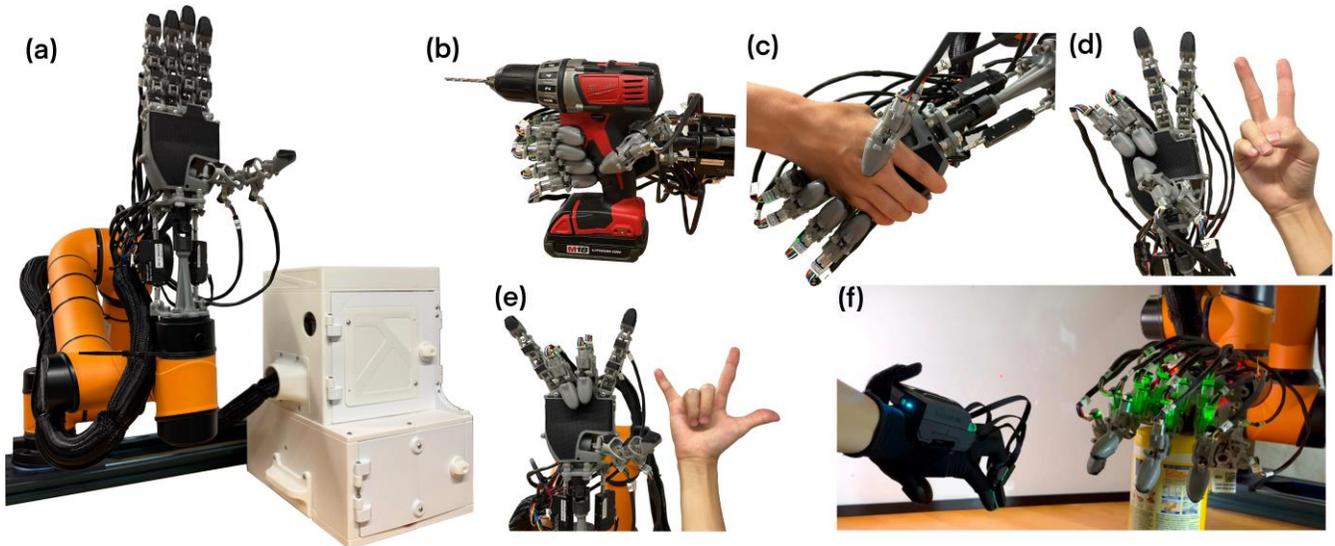

Fig. 1. The developed Krysalis Robotic Hand (a) mounted on a AUBO i5 cobot, with its umbilical cord and the robot control unit (RCU), (b) holding a power drill machine, (c) shaking hand with a human, (d) making victory 'V' sign, (e) making the "I Love You" sign, (f) being teleoperated in real-time.

Not surprisingly, integrating all these requirements into a single design presents a significant challenge. Over the past few years, various robotic end-effectors have been developed, each optimized for specific applications. However, their adoption in industry remains limited [3]. Several factors contribute to this, including limited DoFs or severe underactuation [26], overly complex designs, such as tendon-driven mechanisms that require tedious tendon routing [27], [28], and excessive weight, which confines their use to more costly high-payload industrial or collaborative robots [10], [29].

As one notable past effort, the HRI Hand features 15 joints and a partially anthropomorphic design [30]. However, due to kinematic constraints, its DoF are reduced to six, using a limited number of linear actuators connected to the fingers at the metacarpophalangeal (MCP) joints, along with a separate actuator for the thumb's carpometacarpal (CMC) joint. The HRI Hand also lacks the Radial-Ulnar Deviation (RUD), which is an important DoF in the human wrist. As another example, in a recent patent, published by Tesla, Inc. for its Optimus robot, the robotic hand utilizes a tendon-driven mechanism for the finger actuation [31]. Although the design achieves impressive aesthetics, it uses only six actuators. With the distal and proximal phalanges fused into a single piece, anthropomorphism and dexterity are significantly reduced. Another recent example of a partially anthropomorphic robotic hand is the LEAP Hand [32]. This design employs a direct drive mechanism with all actuators integrated within the fingers and palm. However, it contains only three fingers and a thumb with four DoFs per finger. Despite having one less finger than a human hand, it remains 30% larger, further compromising its anthropomorphism [32]. Moreover, it does not include any DoF in the wrist.

Another notable past design is the Shadow Dexterous Hand by Shadow Robot Company [10]. This hand has 20 DoF, actuated by 20 DC motors connected to the joints through tendons, all placed in the forearm of the hand [33]. By housing the actuators and associated mechanisms in the spacious forearm, this design achieves a relatively thinner palm, which comes at the cost of a bulky actuation unit that increases the weight of the robot to 4.3 kg [10]. Since each joint of the finger is actuated by a tendon, any failure or breakage requires hours of maintenance and downtime. As discussed in [34], independent joint actuation via multiple tendons increases assembly complexity, adversely affects weight, and reduces driving efficiency. The large weight of Shadow Hand has made it incompatible with lower-payload, lower-cost robotic arms [35] and its high price tag has made it unaffordable to many research laboratories, thus limiting its wide adoption [36].

The SCHUNK SVH 5-finger hand (S5FH), designed by SCHUNK, Germany, is yet another significant contribution to the field of dexterous anthropomorphic robotic hands [37]. The S5FH uses a leadscrew actuation mechanism housed within the palm cavity, which helps to reduce its overall form factor [38], [39]. As such, SCHUNK SVH has demonstrated remarkable anthropomorphism and robustness. With 20 joints and several kinematic constraints, it provides 9 degrees of freedom (DoF). This limitation reduces the number of gestures and grasps it can perform [37]. Again, the extremely high price of the hand limits its widespread adoption in academia and industry [40].

Finally, most robotic hands focus primarily on finger actuation, often neglecting the importance of wrist mobility in manipulation [41], [42] and therefore reducing their anthropomorphism and dexterity [34], [37], [43].

To address these challenges and advance research in dexterous robotic manipulation, we introduce the **Krysalis Hand**, a five-finger, 18-DoF, high-payload anthropomorphic robotic hand. Weighing 790 grams, it can support a passive payload of over 10 lbs and exert nearly 10 N of active force per finger. The design achieves several competing requirements, including simplicity, lightweight construction, high DoFs, high range of motion (RoM), anthropomorphism, high payload capacity, large active force, and affordability. With fully actuated fingers and thumb phalanges, the Krysalis Hand conforms to object contours, allowing for secure grasping and dexterous manipulation of various shapes. Moreover, the Krysalis Hand includes a dedicated wrist module with two precision linear actuators across two DoFs, namely flexion-extension (FE) and radial-ulnar deviation (RUD). This design closely mimics the motion of the human wrist, enhancing its anthropomorphic qualities.

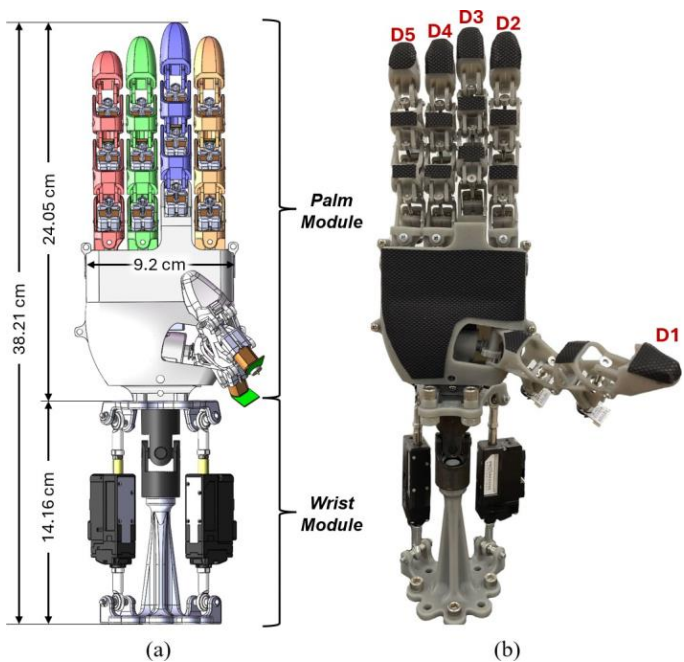

Fig. 2. The Krysalis Hand (a) CAD assembly showing wrist and palm modules, with fingers in different colors (b) designed and implemented robotic hand with finger numbering.

The Krysalis Hand is designed to decouple payload handling from active manipulation forces, addressing both requirements independently. The self-locking mechanism allows the hand to support large passive payloads without continuous motor effort, while each finger can exert 10 N of active force, ensuring a strong and stable grip for effective manipulation. This approach reduces motor torque demands, making the system lighter, more efficient, and cost-effective.

## II. SPECIFICATIONS

The design of the Krysalis Hand reflects that of the human hand (Fig. 2) with four fingers, a thumb, and a wrist with each of the phalanges and joints independently actuated. Table I lists the specifications of the Krysalis Hand.

## III. MECHANICAL DESIGN

The human hand is a marvel of evolution, capable of generating large forces for labor-intensive tasks such as squeezing and gripping, while also supporting large payloads. Simultaneously, it enables highly dexterous coordination and manipulation across a wide range of tasks with various speed and force requirements, from precise object assembly requiring large active forces to tying knots or playing musical instruments. In contrast, most electric motors used in robotics, such as brushed and brushless DC motors, have a limited range of optimal performance. Although they can achieve reasonable peak power under specific speed and torque conditions, they struggle to handle the diverse set of tasks that human hands perform [44]. At low speeds, motors become inefficient and generate excessive heat, leading to potential damage, and at high speeds, they fail to produce sufficient torque.

The Krysalis Hand integrates compact electric motors directly into the dorsal side of the fingers, paired with optimized gear ratios to deliver practical torque output across a broad range of operating speeds. This allows the robot to perform versatile manipulative tasks while maintaining a form factor that is significantly lighter and more compact than the state-of-the-art designs.

TABLE I
KRYSALIS HAND: SPECIFICATIONS

| | |
|---|---|
| Weight of the Hand without Wrist | 470 g (1.04 lbs) |
| Weight of the Hand with wrist | 790 g (1.74 lbs) |
| Actuators | 16 Servo Motors, 2 Linear Actuators |
| Operating Voltage (V) | 6 - 12 V |
| Degrees of Freedom (DoFs) | 18 (16 in Hand & 2 in Wrist) |
| Joints | 20 |
| Finger Speed (°/s) | 91.5°/s |
| Passive Payload Capacity | > 4.53 kg (10 lbs) |
| Active Finger Force | 10 N |

### A. Leadscrew Finger Actuation Mechanism

To the best of our knowledge, no current robotic hand, in either the commercial market or scientific community, offers fully actuated, non-tendon-driven fingers for all five fingers. Previous designs predominantly rely on a single actuator to control all three phalangeal joints of a finger, namely, the distal interphalangeal (DIP), proximal interphalangeal (PIP), and metacarpophalangeal (MCP) joints for the second through fourth digits (D2–D4), as well as the interphalangeal (IP) and carpometacarpal (CMC) joints for the first digit (D1) [45], [46]. These designs typically incorporate either linear actuators in the palm, driving the metacarpophalangeal (MCP) joint via a linkage mechanism that underactuates the proximal (PIP) and distal (DIP) joints, or a motor in the palm that employs tendons to actuate all three joints simultaneously. Such kinematic constraints prevent these finger mechanisms from fully replicating the capabilities of the human hand, resulting in reduced grasping force and dexterity [47]–[49].

The Krysalis Hand utilizes a motor-driven leadscrew mechanism to power each phalange joint. Despite increasing the number of required actuators, the design maintains a compact footprint and an anthropomorphic palm. It also provides a high gear ratio and, by generating large active forces across a wide range of working conditions, enables the execution of diverse grasp tasks. Also, as discussed in section III-B, this mechanism leverages self-locking, allowing the system to maintain its position without active motor input, making it especially beneficial for handling large payloads over extended durations. Fig. 3a demonstrates the finger kinematics, where the same mechanism repeats for all joints (DIP, PIP, and MCP). Fig. 3b presents a detailed close-up of the DIP joint, illustrating the leadscrew actuation mechanism and its components.

designs are prone to back-drivability, where external forces cause unintended joint rotation, complicating manipulation and control, while increasing power consumption, leading to overheating and reduced motor lifespan.

To overcome these limitations, in the Krysalis hand, we leverage the self-locking capability of the leadscrews. In this approach, the finger is actuated to the desired kinematics, where it can hold position regardless of the external forces. This characteristic simplifies control during manipulation and increases payload capacity, which is theoretically only limited by the mechanical strength of the hand.

The schematic of Fig. 4 depicts the lead angle ($\alpha$), lead ($l$), mean diameter ($d$), and friction angle ($\phi_s$). The lead angle is a measure of steepness of the screw helix. The friction angle ($\phi_s$) is defined as the angle between the resultant force $R$, formed by the normal force, $N$, and the static friction force, $F$, for a given coefficient of friction $\mu$ between the lead screw and the nut. The self-locking property is achieved when $\phi_s > \alpha$ [51].

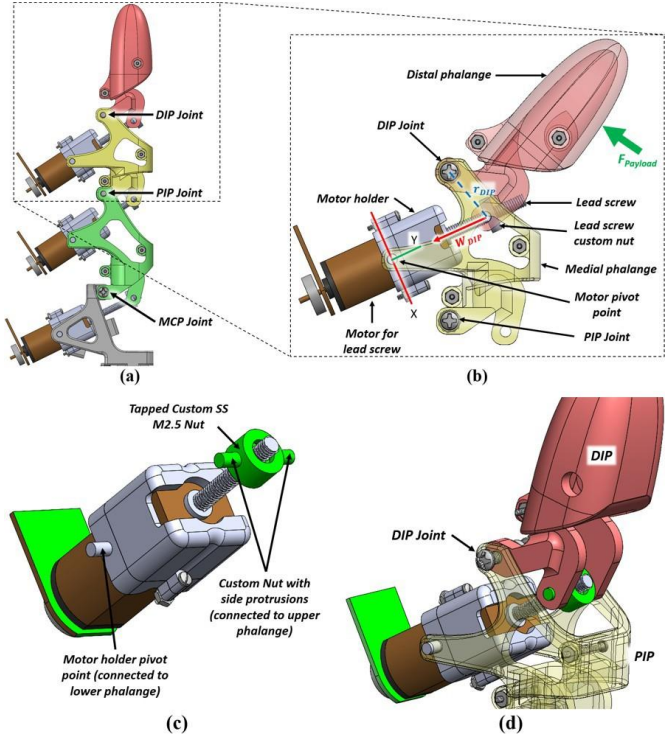

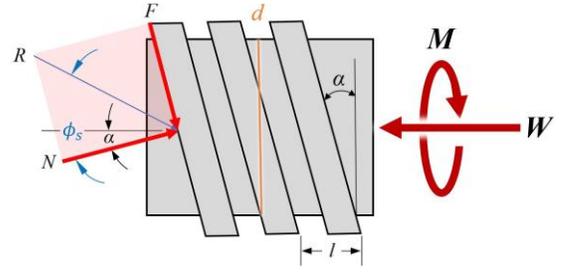

Fig. 4. Free body diagram of the screw.

Fig. 3. (a) Entire finger assembly, (b) Leadscrew actuation mechanism in detail, (c) Schematic of motor and custom nut in isolation, (d) Schematic showing the custom nut engaged to the phalange above it (DIP in this case).

The actuation mechanism includes a custom leadscrew, which also serves as the output shaft of an electric motor, driving a custom stainless steel M2.5 nut that moves along its axis (Fig. 3c). This custom nut features pins on both sides that insert into the pivot point of the phalange above it, enabling it to engage with the finger and rotate relative to it. This forms part of a two-link rocker mechanism, where one of the links has a variable length (Fig. 3d). Rotation of the leadscrew creates an axial force ($W_{DIP}$) on the nut, as shown in Fig. 3b, which subsequently generates a torque around the DIP joint. The torque moment arm is the distance between the phalange joint and the center of the leadscrew nut (denoted as $r_{DIP}$ in Fig 3b). This torque balances any active resistance force at the fingertip. The same mechanism is used in all other phalanges of the D2-D5 fingers.

### B. Self-Locking

In most robotic hand designs, motors must exert torque continuously during gripping and manipulation [50]. Such

In Krysalis Hand, the lead screw incorporated in the motor gearbox is a M2.5 screw with 16 mm length for the DIP and PIP joints and 20 mm length for the MCP joint. Two different leadscrew lengths have been used to maximize RoM and hence dexterity. The lead ($l$) of the screw is 0.35 mm. The leadscrew mean diameter ($d$) is 2.50 mm. The coefficient of friction between stainless steel leadscrew and stainless-steel nut (dry) is $\mu \approx 0.42$ [52]. The following equation provides the lead angle ($\alpha$) [51]:

$$\alpha = tan^{-1}\left(\frac{l}{\pi d}\right) \Rightarrow \alpha = \mathbf{2.55°} \quad (1)$$

Also from Fig. 4 friction angle ($\phi_s$) can be calculated as:

$$\phi_s = tan^{-1}\left(\frac{F}{N}\right) \Rightarrow \phi_s = tan^{-1}(\mu) \Rightarrow \phi_s \approx \mathbf{21.83°} \quad (2)$$

Comparing the results in Eqs. 1 and 2, it is noted that the friction angle ($\phi_s$) > lead angle ($\alpha$). As such, self-locking is ensured.

### C. Thumb Carpometacarpal (CMC) Joint

The significance of the thumb (D1) in enabling various types of grasp cannot be overstated. This is especially true for power and precision grasps, which rely on D1 and the rest of the fingers to exert force on the payload [53]. The unique opposing orientation of the thumb enables a wide range of gripping and manipulation techniques; without it, even the simplest tasks would be impossible.

The thumb also has three joints: Interphalangeal (IP), Metacarpophalangeal (MCP), and Carpometacarpal (CMC), named from the fingertip to the base. It is important to note that D1's joints differ slightly from those of the other four digits. Unlike D2-D5, where all joints flex and extend in the same plane, the thumb's CMC joint operates in a different plane, distinct from IP and MCP. This unique joint enables us to grasp large objects while still allowing the IP joint to touch all other fingertips. This feature has been implemented in the Krysalis Hand by adopting a different type of actuation mechanism for the CMC joint.

By utilizing the available space in the palm cavity, a worm gear set has been designed for the CMC. Powered by a high-speed DC motor, the driver worm is coupled to the motor. The driver worm is supported between the motor and the inner wall of the palm using a ball bearing. This driver worm rotates the worm wheel (which is designed to be a part of the CMC link), allowing the thumb to achieve powerful grasps and high speeds. Fig. 5 shows the thumb CMC actuation mechanism.

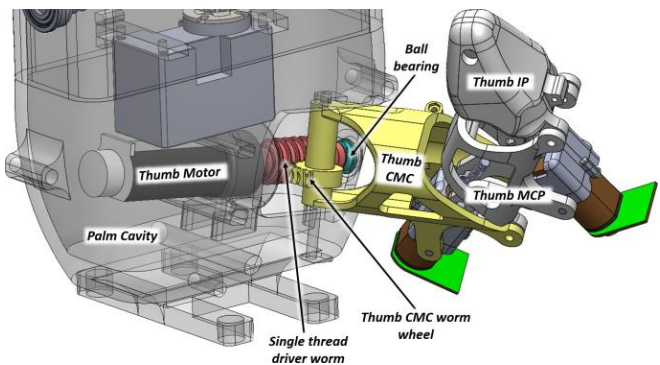

Fig. 5. Thumb CMC joint mechanism.

The Krysalis Hand is designed to hold a payload in place without motor use, conserving energy and reducing wear. The thumb, responsible for balancing the forces of the four fingers, also requires a self-locking CMC joint.

The lead of the worm thread, $l_{worm} = 2.5$ mm, and the pitch diameter, $d_{worm} = 9.49$ mm. Plugging these into Eq. 1, we have: $\alpha_{CMC} = 4.80°$. Next, the friction angle of the worm gear at the thumb CMC is obtained by Eq. 2, assuming the coefficient of friction of the 3D printed aluminum worm/wheel gear pair to be $\mu_{CMC} = 0.46$ [54]. Plugging in these values, we get $\phi_{s,CMC} = 24.7°$. As such, for the thumb CMC we have: $\phi_{s,CMC} > \alpha_{CMC}$, achieving self-locking.

Hence, the payload limit of the Krysalis Hand is determined by its mechanical strength, specifically the load at which structural damage is likely to occur.

### D. Abduction–Adduction Mechanism

For grasping objects larger than the width of the human hand or performing complex manipulations, abduction-adduction plays a crucial role [55]. Incorporating this degree of freedom, which lies in a plane orthogonal to the flexion-extension of the fingers, presents significant challenges. This is especially true for tendon-driven robots, where the limited space makes routing multiple tendons difficult [27].

The space in the palm of the Krysalis Hand allows for actuator integration, providing sufficient room for the abduction-adduction mechanism for D2-D5 fingers. However, studies indicate that abduction-adduction movements tend to occur in tandem during tasks such as manipulation, power, or precision grips, with little added benefit from independent abduction-adduction control for each finger [56]. To reduce complexity, cost, and weight, we chose to couple abduction-adduction movements across D2 to D4 using an actuation mechanism powered by one servo motor. To achieve such synchronous movements, a gear train has been developed. The schematic in Fig. 6 demonstrates this design. Although not as critical for abduction-adduction, the self-locking feature of this DoF can also be demonstrated for the Krysalis Hand.

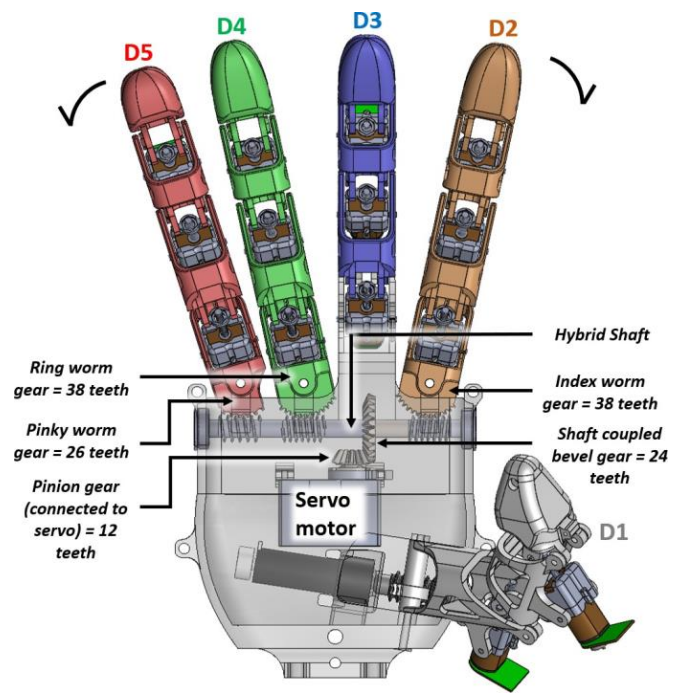

Fig. 6. Abduction–adduction mechanism.

In this design, a servo motor, embedded within the palm, is coupled to a pinion (with 12 teeth), which meshes with a bevel gear (24 teeth). The bevel gear is part of a hybrid transmission shaft that simultaneously drives 3 worm gears. Each worm on the transmission shaft engages with a worm wheel located at the corresponding finger root. The worm for D2 is left-handed, while the worms for D4 and D5 are right-handed, accommodating their opposite movements. The worm pitch for D2 and D4 is 2 mm, while D5 has a pitch of 2.63 mm. This allows D5 to move faster than D2 and D4, thus achieving a larger span in the same duration, similar to the human hand. The third digit (D3) is excluded from the abduction-adduction actuation mechanism to maintain a symmetric grasp configuration. It is important to note that the number of DoF of the Krysalis Hand can be easily extended to 21 simply by accommodating 4 motors at the back of the hand or within the palm for independent abduction-adduction of each finger.

### E. The Wrist Module

The wrist of the human hand is another anatomical marvel. It has two primary DoFs: flexion-extension (FE) and radial-ulnar deviation (RUD), occurring at the same joint but in perpendicular planes. Flexion involves bending the wrist forward toward the palm, and extension involves bending it backward toward the dorsal side of the hand (this actuation takes place in the same plane as the finger flexion-extension). RUD enables side-to-side movement. The radial deviation moves the wrist toward D1, while the ulnar deviation moves it toward D5 (this actuation takes place in the same plane as the finger abduction-adduction). Although some may consider forearm pronation-supination (PS) (rotation of the hand about the forearm axis) a third wrist DoF, it occurs primarily at the elbow rather than within the wrist itself [57]. Furthermore, PS can be easily accommodated on the last link of almost all robotic arms and hence does not need to be included in the hand.

To closely emulate the human wrist motion, a 2-DoF motion platform has been designed for the Krysalis Hand. This platform uses two linear actuators, each with a stroke length of 26.92 mm (1.06 in) and a maximum dynamic force of 100 N (22.48 lbs). Together, they support the wrist of the Krysalis Hand. These two commercially available self-locking linear actuators connect the upper platform (mounting the palm module) and the lower wrist platform (see Fig. 7) through fisheye (spherical) joints at both ends. These spherical joints are positioned 90° apart on the dorsal side of the hand (Fig. 8a). Additionally, the upper and lower mounting plates are linked by a 90° universal joint, designed to perform FE and RUD DoFs at a single point.

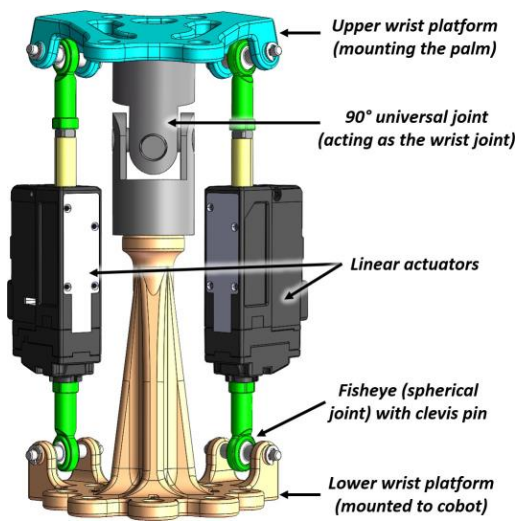

Fig. 7. The mechanism of the wrist module.

The fisheye spherical joint's maximum swivel angle is approximately 40°. As such, to maximize wrist FE and RUD range, the upper wrist platform was designed with a 30° angle between the horizon and the spherical joint axis, shown in Fig. 8b. In this configuration, the wrist achieves a palmar flexion of 52°, dorsal extension of 18°, and radial-ulnar deviation of 18°.

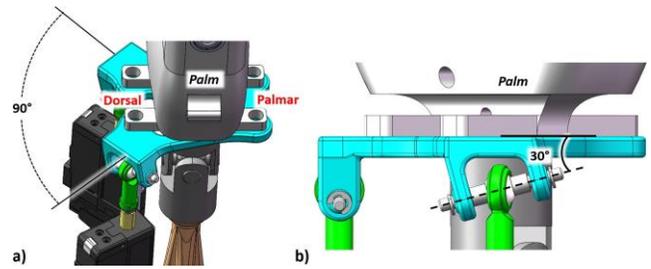

Fig. 8. The design of the upper wrist platform with (a) 90° offset between the two linear actuators and (b) 30° angle between the horizon and the spherical joint axis.

## IV. MATERIALS AND FABRICATION PROCESS

The fabrication of the majority of the components for the Krysalis Hand was carried out using stereolithography (SLA) 3D printing. For most parts, we used Formlabs Form 3B+ printer [58] with Tough 1500 resin for its higher biocompatibility and strain springback capabilities [59]. SLA offers a fine resolution, as small as 25 microns [58]. This level of detail helps with tight tolerances and enables kinematic repeatability. Additionally, SLA 3D printing produces parts with improved isotropic properties [60]–[63], whereas traditional Fused Deposition Modeling (FDM) introduces anisotropy, which can compromise structural stability depending on the loading direction. A subset of parts, such as D1's carpometacarpal (CMC) phalange and its worm gear, were metal 3D printed using Selective Laser Melting (SLM) of AlSi10Mg aluminum alloy to improve reliability. Lastly, the custom nut piece for the finger leadscrew actuation mechanism was machined out of stainless steel.

## V. ELECTRICAL DESIGN

Each phalange in the fingers D1 to D5 (excluding the CMC joint of the thumb) is driven by a micro DC motor with with embedded encoder. Together, these motors provide 14 degrees of freedom for the hand. These 14 micro DC motors/encoders, operating at 6 V, are managed by seven AT8236 2-channel motor controllers, each independently controlling and receiving feedback from two motors. All drivers are supplied through a single power source and communicate with four Arduino Mega microcontrollers, allowing for independent motor control. The encoders in these micro DC motors allow the Arduino Mega microcontroller to interpret the finger position in real time, forming a closed-loop control.

As discussed in section III-C, thumb's CMC joint follows a unique trajectory and occasionally bears the full payload weight. Therefore, it is equipped with a stronger DC motor. This motor is placed in the palm and operates on a separate 12 V power supply. The CMC motor is also controlled by an Arduino Mega, which manages the thumb's MCP and IP joints, using a separate AT8236 2-channel motor controller. The CMC joint provides one additional DoF.

The abduction-adduction finger actuation mechanism is driven by a servo motor, which is housed in the palm cavity. This servo motor is controlled by a Robotis OpenRB-150 module, which communicates using the Transistor-Transistor Logic (TTL) protocol.

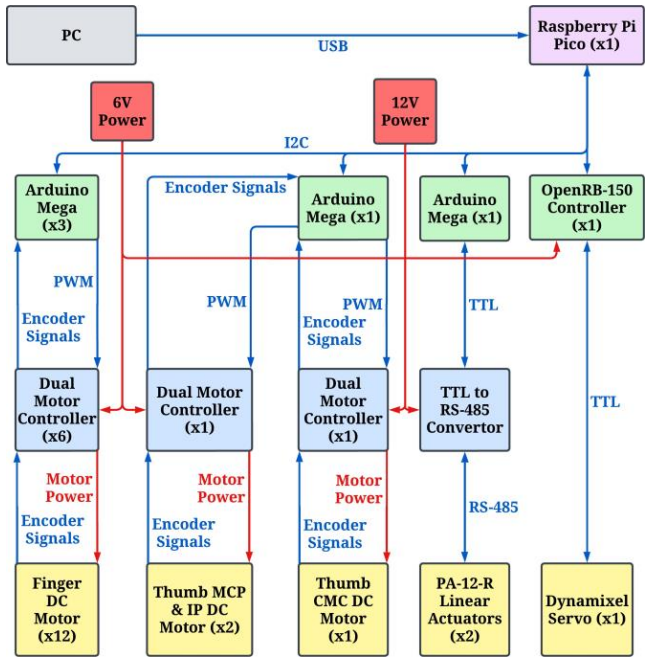

Fig. 9. Electrical block diagram of the Krysalis Hand

The wrist of the Krysalis Hand is driven by two Progressive Automation PA-12-R linear actuators, which operate at 12 V. These linear actuators use TTL to RS-485 communication adapters to communicate with a separate Arduino.

All of the operational modules (shown in green in Fig. 9) are controlled by a separate master Raspberry Pi Pico (shown in purple). Communication between the master Raspberry Pi Pico and motor controllers is handled via a bidirectional I²C central bus.

## VI. EXPERIMENTS AND PROCEDURES

### A. Finger Force

Active finger force refers to the force that the Krysalis Hand can exert during active motor operation. To measure the active fingertip force, KUNWEI KWR75B force sensor was employed in an experimental setup where the finger was pressed against the load cell. As illustrated in Fig. 10, the finger can provide and maintain a maximum active force of nearly 10 N.

Passive force is the maximum load that the Krysalis Hand can hold statically in a fixed position without requiring continuous motor actuation. Thanks to its self-locking mechanism, the passive force capacity is ultimately limited by the mechanical strength of the hand. This mechanical strength is dictated primarily by the design and material selection.

To minimize stress concentration and maximize passive force handling, we performed finite element analysis (FEA). The current design utilizes SLA 3D printing for rapid prototyping of most of the components. Replacing all parts with alternative materials, such as metal, could significantly improve Krysalis Hand's passive load capacity. To avoid potential damage during passive load testing, we imposed a payload limit of 10 lbs (Fig. 11). However, FEA results indicate that component failure occurs under significantly higher loading conditions, suggesting that the actual passive load capacity is greater, even with the current material.

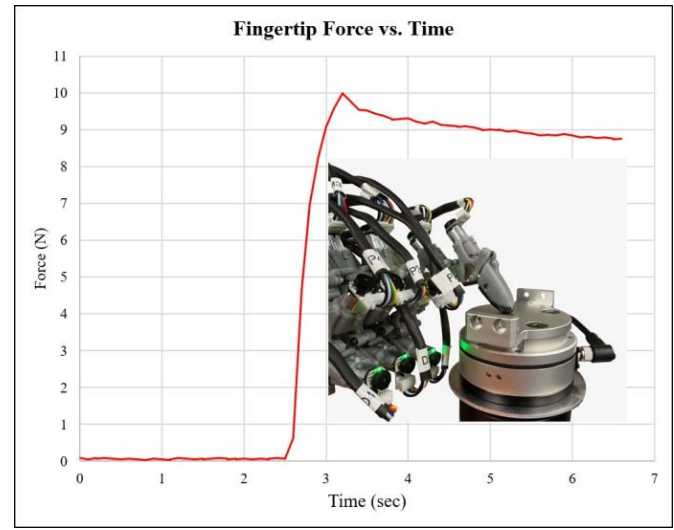

Fig. 10. Active fingertip force as a function of time.

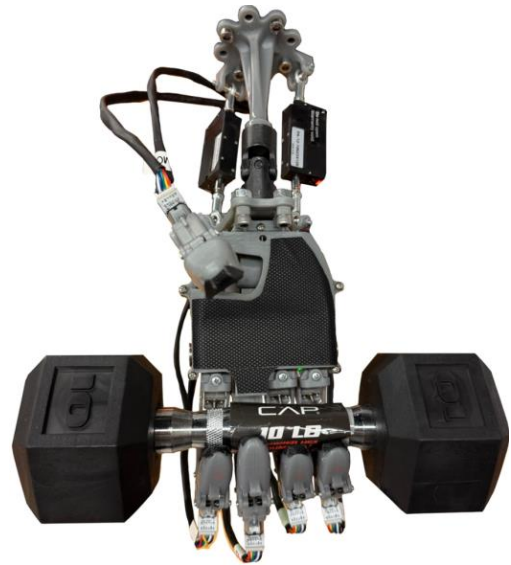

Fig. 11. Krysalis Hand passively holding a 10 lbs (4.54 kgs.) dumbbell.

### B. Thumb Opposability and Dexterity

To evaluate dexterity, thumb opposition was tested, which is the ability of the thumb to touch the tips of each of the four fingers [53], [64]. The test was carried out by moving the thumb (D1) to sequentially meet the tips of D2, D3, D4, and D5, respectively. Each interaction was analyzed based on: (1) accuracy in reaching the fingertip position, (2) repeatability, and (3) complete fingertip-to-thumb contact without gaps (contact success).

The experiment was carried out multiple times for each finger. The thumb achieved contact success in all repetitions, which was visually verified. Example images of thumb opposition for each finger are shown in Fig. 12.

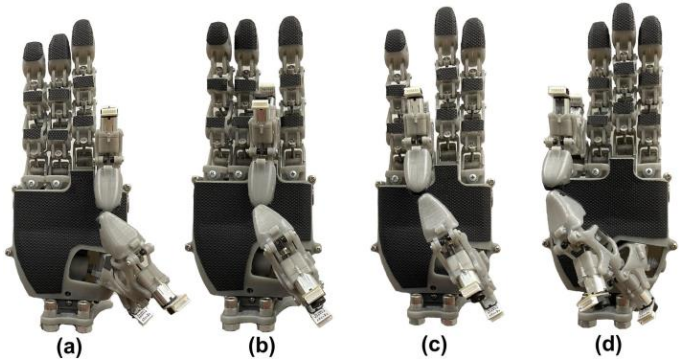

Fig. 12. Thumb opposition experiment with (a) D2, (b) D3 (c) D4, and (d) D5.

To further evaluate the Krysalis Hand, the workspaces of the fingers and thumb (representing the entire palm module, excluding the wrist module) are explored and the trajectories associated with all fingertips are recorded. As illustrated in Fig. 13, these trajectories reveal the spatial extent of each digit's motion. Notably, the regions where the workspaces of the four fingers intersect with that of the thumb confirm effective thumb opposability, which is a critical feature for enabling precision and power grasps.

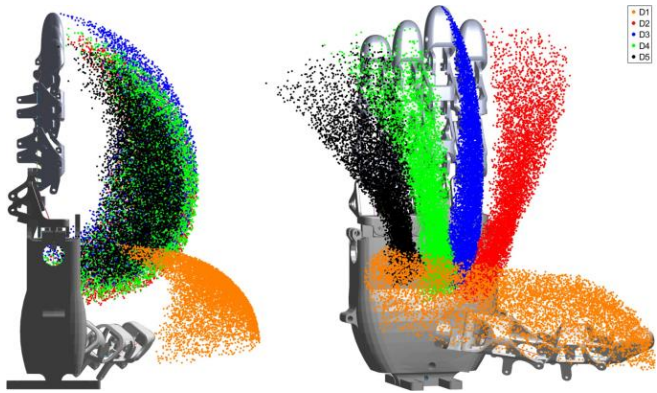

Fig. 13. Generated workspace for the five fingers (palm module) showing the thumb opposability regions.

### C. Joint Range of Motion

To evaluate the RoM of the Krysalis Hand, it has been benchmarked against the documented RoM of the human hand [65]–[68] as shown in Table II. The results indicate that the total RoM across all five fingers of the Krysalis Hand surpasses that of the average human hand, demonstrating enhanced dexterity in individual finger joints. It achieves smaller ROM for the wrist, particularly in extension. On average, the Krysalis Hand achieves a 11.2% greater range of motion (RoM) compared to the human hand.

TABLE II
COMPARISON OF ROM BETWEEN KRYSALIS HAND AND HUMAN HAND FOR CMC [65], MCP, IP, PIP, DIP [66], AB/ADDUCTION [67], AND WRIST [68] JOINTS.

| Part | Joint/Category | Krysalis Hand (°) | Human Hand (°) |
|---|---|---|---|
| Thumb | CMC | 106.24 | 55.00 |
| | MCP | 52.72 | 57.27 |
| | IP | 45.02 | 65.00 |
| | **Total RoM** | **203.98** | **177.27** |
| Index | MCP | 103.13 | 49.20 |
| | PIP | 75.07 | 86.60 |
| | DIP | 68.09 | 57.95 |
| | Ab/adduction | 26.73 | 19.19 |
| | **Total RoM** | **273.02** | **212.94** |
| Middle | MCP | 101.92 | 66.33 |
| | PIP | 73.46 | 85.08 |
| | DIP | 73.04 | 55.62 |
| | **Total RoM** | **248.42** | **207.03** |
| Ring | MCP | 100.56 | 65.30 |
| | PIP | 72.93 | 93.67 |
| | DIP | 73.57 | 58.43 |
| | Ab/adduction | 26.73 | 17.29 |
| | **Total RoM** | **273.79** | **234.69** |
| Pinky | MCP | 98.93 | 52.76 |
| | PIP | 72.03 | 91.81 |
| | DIP | 72.05 | 56.71 |
| | Ab/adduction | 39.37 | 45.01 |
| | **Total RoM** | **282.38** | **246.29** |
| Wrist | Flexion | 52.00 | 60.00 |
| | Extension | 18.00 | 60.00 |
| | Radial deviation | 18.00 | 20.00 |
| | Ulnar deviation | 18.00 | 30.00 |
| | **Total RoM** | **106.00** | **170.00** |

### D. Grasping Capability Assessment

The Krysalis Hand was evaluated using various standard grasp types from the GRASP Taxonomy [69]. These tests involved handling objects of various shapes, sizes, weights, and surface textures, including a hex nut, tape, card, clip, key, cylindrical container, scissors, and smartphone. The selected items were chosen to represent both industrial objects, such as tools, and everyday items like personal belongings and household objects, ensuring the evaluation reflects a broad spectrum of real-world use cases. We categorized the results into three grasp types—precision grasp, power grasp, and tripod grasp—as depicted in Fig. 14. These classifications were chosen to highlight the versatility of the Krysalis Hand in performing tasks that require different levels of dexterity, grip force, and contact stability. While the results are organized within these three taxonomies for simplicity and clarity, the Krysalis Hand is not confined to them.

### E. Anthropomorphism

The design of the Krysalis Hand was guided by the US Army Hand Anthropometry Data [70] to ensure an anthropomorphic form. This dataset, based on measurements from 1,003 men and 1,304 women, helped define key specifications such as size and proportions, ensuring that the final design closely resembles a typical human hand. Fig. 1c-e compares the Krysalis Hand to a human hand. Integrating the finger actuation mechanism within the inner lining of the fingers and accommodating the abduction-adduction mechanism resulted in minor dimensional deviations from an average human hand.

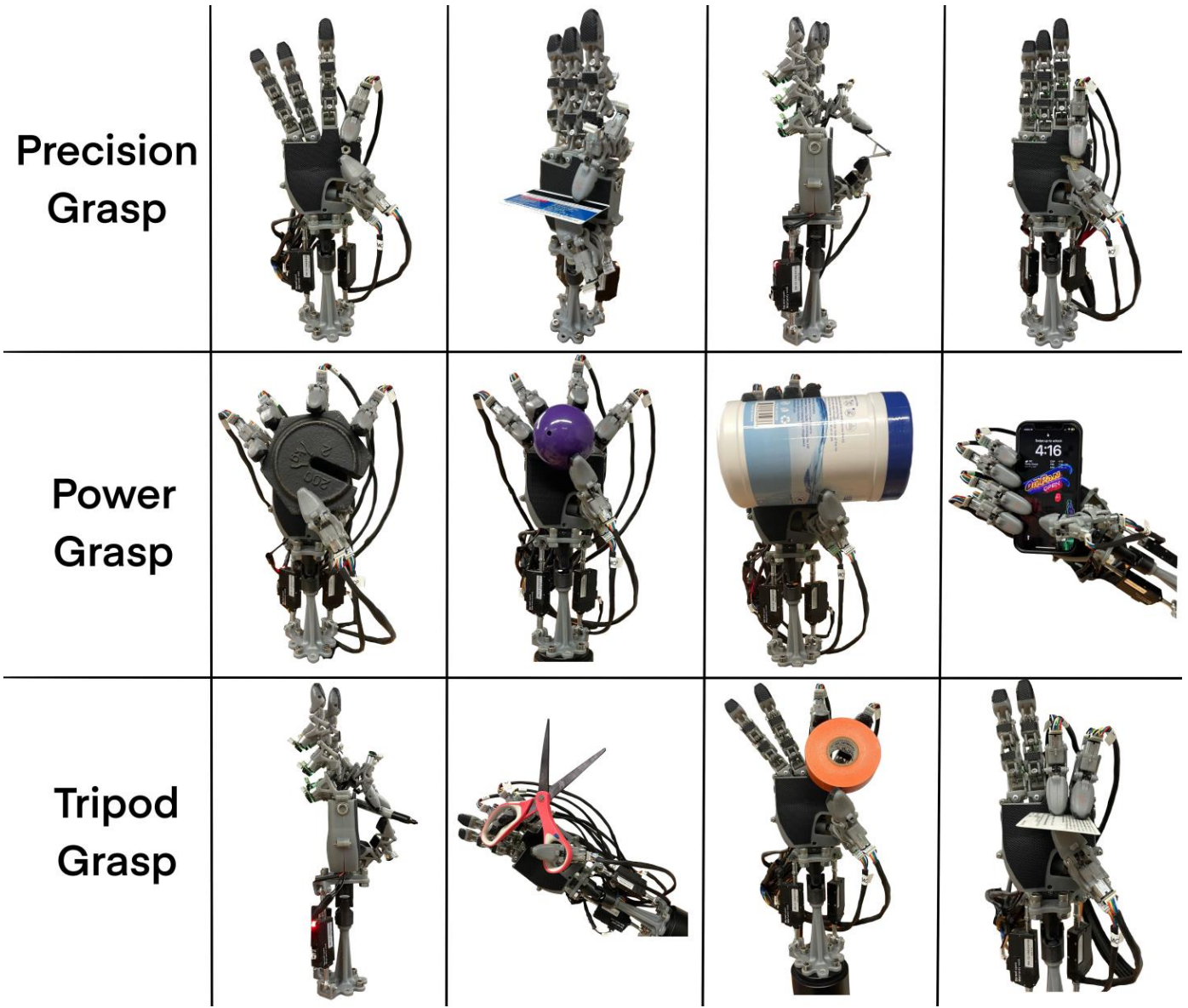

Fig. 14. The grasping capabilities of the Krysalis Hand with various payloads, categorized into precision, power, and tripod grasps.

Table III presents a comparison of the anthropometric data of various parts of the human hand and the corresponding dimensions of the Krysalis Hand components.

From the table, it is evident that the Krysalis Hand exceeds the average length of human hand components (derived from anthropometric data) by 24.34%. In contrast, the widths of these components are reduced by 7.15% compared to the human hand. Furthermore, based on anthropometric data from [70] and the values presented in Table III, the average palm-facing surface area of the human hand is estimated at 219.4 cm², while that of the Krysalis Hand is 242.6 cm². This makes the Krysalis Hand approximately 10.6% larger in surface area than the average human hand. Notably, this increase was achieved while integrating all actuators within the palm cavity and along the dorsal side of the fingers, maintaining a lightweight design, and accommodating a high number of degrees of freedom.

*F. Teleoperation*

High-precision teleoperation is essential for tasks involving dexterous manipulation [71], [72]. This is especially relevant in learning-from-demonstration frameworks, where accurate reproduction of complex hand movements, such as those required in fine assembly tasks on production floors, is critical. While recent advances in visual hand tracking using VR headsets and other vision-based methods have shown promise, their precision remains insufficient for high-fidelity demonstration or teleoperation. Therefore, in this work, we adopt a hardware-based approach using MANUS motion-capture gloves to measure detailed finger motion [73]. These gloves were selected for their high spatial and temporal resolution, low-latency wireless communication, and compatibility with the multi-DoF architecture of the Krysalis Hand. To implement teleoperation between the MANUS glove and the Krysalis Hand, a ROS 2 teleoperation package was developed. This package is a modified version of the approach described in [71]. The ROS 2 package captures the DIP and fingertip Cartesian coordinates for each finger via the MANUS glove and maps them to the Krysalis Hand's DIP and fingertip positions on the developed URDF in the PyBullet environment (Fig. 15a). The package utilizes PyBullet's inverse kinematics function to calculate the angles for each joint on the Krysalis Hand (Fig. 15b). These angles are then broadcast via ROS 2 messages to the master Raspberry Pi Pico that controls all of the slave Arduinos (Fig. 9).

TABLE III
DIMENSIONS OF VARIOUS PARTS OF THE HUMAN HAND [70] AND THE CORRESPONDING KRYSALIS HAND COMPONENTS

| Hand Part | Lengths (cm) | | Widths (cm) | |
|---|---|---|---|---|
| | Human Hand | Krysalis Hand | Human Hand | Krysalis Hand |
| Palm | 11.07 | 12.90 | 9.53 | 9.20 |
| Index Finger | 10.83 | 13.01 | 2.30 | 1.90 |
| Middle Finger | 8.38 | 10.25 | 2.25 | 1.90 |
| Ring Finger | 10.69 | 13.09 | 2.14 | 1.90 |
| Pinky Finger | 8.60 | 12.20 | 1.92 | 1.90 |
| Thumb | 12.34 | 12.23 | 2.40 | 2.59 |
| Wrist | 0.00 | 3.30 | 6.58 | 5.79 |
| **Percent Change** | **+24.34%** | | **-7.15%** | |

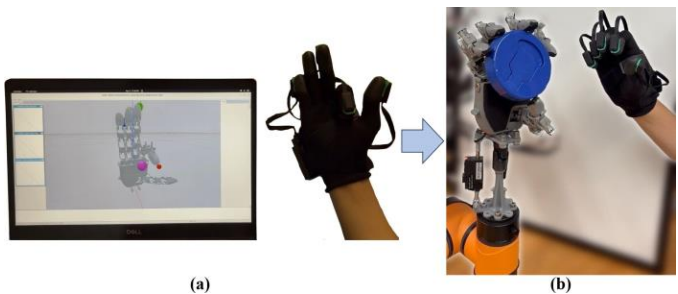

Fig. 15. Teleoperation setup showing real-time (a) URDF control in the PyBullet environment (b), which broadcasts the ROS 2 messages to the Krysalis Hand.

We utilized MicroROS, a lightweight library that brings ROS 2 functionality to microcontrollers such as the Raspberry Pi Pico [74]. Specifically, it enabled the Raspberry Pi Pico to act as a ROS 2 node capable of subscribing to ROS 2 messages over a serial or UDP transport layer. This allowed integration with the ROS 2 ecosystem, facilitating real-time communication and control between the Krysalis Hand's embedded hardware and external ROS-based system.

## VII. CONCLUSION

This paper presents the detailed design of the Krysalis Hand: a compact, lightweight, high-payload, high-degree-of-freedom, anthropomorphic, five-finger robotic end-effector. The Krysalis Hand integrates all actuators within the hand while maintaining compact size and functionally anthropomorphic design. Being only 10.6% larger than the average human hand, the Krysalis Hand offers 18 DoF. It weighs just 790 grams, making it a practical choice for robotic end-effectors in both industrial and everyday applications. Its self-locking mechanisms enable it to handle heavy payloads, limited only by its material strength. The experiments demonstrated a maximum active finger push force of nearly 10 N, ensuring firm grips and secure handling of heavy objects. Additionally, results demonstrated thumb opposability, and an average 11.2% greater range of motion compared to the human hand. We have showcased the grip stability of the Krysalis Hand across diverse grasping tasks. Finally, the Krysalis Hand was teleoperated using the motion-capture technique via the MANUS glove.

## VIII. Biography Section

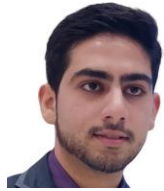

**Al Arsh Basheer** received his B.S. in Mechanical and Aerospace Engineering from the University at Buffalo, State University of New York, USA in 2021. Later, he obtained his M.S. in Robotics in 2023 from the University of California, Davis, USA. His research focuses on developing high-dexterity humanoid robots to innovate automation across electronics and aerospace industries.

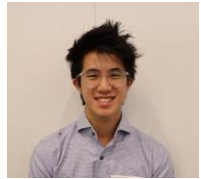

**Justin Chang** is a current 3rd-year undergraduate student working towards a B.S. in computer science and engineering from the University of California, Davis. He expects to graduate in 2026. His research interests include robotics and embedded systems.

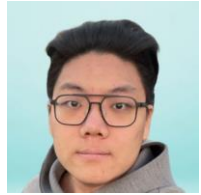

**Yuyang Chen** earned a B.S. in Mechanical Engineering from the University of California, Davis, in 2024 and is currently pursuing an M.S. in Robotics at Johns Hopkins University. His research focuses on bionic robots, humanoid robots, and exoskeletons.

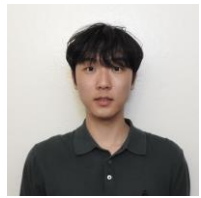

**David Kim** is currently a 3rd-year undergraduate student working towards a B.S. degree in Computer Science and a minor in Mathematics from the University of California, Davis. He is expected to graduate in Spring, 2026. He plans to pursue a PhD in Computer Science following his undergraduate studies. His current research interests include machine learning and computational geometry.

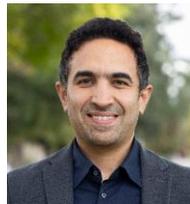

**Iman Soltani** is an assistant professor of Mechanical and Aerospace Engineering and a faculty member in graduate groups of the Departments of Computer Science and Electrical and Computer Engineering at the University of California, Davis. His research spans the interface of artificial intelligence, instrumentation, controls, and design, with a focus on machine learning and robotic systems.

Before joining UC Davis, he worked at Ford Greenfield Labs in Palo Alto, CA, where he founded and led the Advanced Automation Laboratory. He earned his bachelor's, master's, and PhD in mechanical engineering from Tehran Polytechnic (Iran), the University of Ottawa (Canada), and the Massachusetts Institute of Technology (MIT), respectively. He holds more than 18 patents and has authored over 40 journal and conference publications on topics ranging from medical imaging to autonomous driving, high-speed nanorobotics, dexterous bimanual robotics, machinery health monitoring, and precision positioning systems. His research has been featured in prominent outlets such as The Boston Globe, Elsevier Materials Today, ScienceDaily, and MIT News. Among his numerous awards are the MIT Carl G. Sontheimer Award and National Instruments' Engineering Impact Award.